# The Greedy Miser: Learning under Test-time Budgets

**Zhixiang (Eddie) Xu**                                                     XUZX@CSE.WUSTL.EDU
**Kilian Q. Weinberger**                                                      KILIAN@WUSTL.EDU
Washington University, St. Louis, MO 63130, USA

**Olivier Chapelle**                                                        OLIVIER@CHAPELLE.CC
Criteo, 411 High Street, Palo Alto, CA 94301, USA

## Abstract

As machine learning algorithms enter applications in industrial settings, there is increased interest in controlling their cpu-time during testing. The cpu-time consists of the running time of the algorithm and the extraction time of the features. The latter can vary drastically when the feature set is diverse. In this paper, we propose an algorithm, *the Greedy Miser*, that incorporates the feature extraction cost during training to explicitly minimize the cpu-time during testing. The algorithm is a straightforward extension of stage-wise regression and is equally suitable for regression or multi-class classification. Compared to prior work, it is significantly more cost-effective and scales to larger data sets.

## 1. Introduction

The past decade has witnessed how the field of machine learning has established itself as a necessary component in several multi-billion-dollar industries. The applications range from web-search engines (Zheng et al., 2008), over product recommendation (Fleck et al., 1996), to email and web spam filtering (Weinberger et al., 2009). The real-world industrial setting introduces an interesting new problem to machine learning research: computational resources must be budgeted and costs must be strictly accounted for *during test-time*. Imagine an algorithm that is executed 10 million times per day. If a new feature improves the accuracy by 3%, but also increases the running time by $1s$ per execution, that would require the project manager to purchase 58 days of additional cpu time per day.

At its core, this problem is an inherent *tradeoff* between accuracy and test-time computation. The test-time computation consists of two components: 1. the actual running time of the algorithm; 2. the time required for feature extraction.

In this paper, we propose a novel algorithm that makes this trade-off explicit and considers the feature extraction cost during training in order to minimize cpu usage during test-time. We first state the (non-continuous) global objective which explicitly trades off feature cost and accuracy, and relax it into a continuous loss function. Subsequently, we derive an update rule that shows the resulting loss lends itself naturally to greedy optimization with stage-wise regression (Friedman, 2001).

While algorithms such as (Viola & Jones, 2002) directly attack the problem of fast evaluation for visual object detection, in most machine learning application domains, such as web-search ranking or email-spam filtering, analysis and algorithms for on-demand feature-cost amortization are still in their early stages.

Different from previous approaches (Lefakis & Fleuret, 2010; Saberian & Vasconcelos, 2010; Pujara et al., 2011; Chen et al., 2012), our algorithm does not build cascades of classifiers. Instead, the cost/accuracy tradeoff is pushed into the training and selection of the weak classifiers. The resulting learning algorithm is much simpler than any prior work, as it is a variant of regular stage-wise regression, and yet leads to superior test-time performance. We evaluate our algorithm's efficacy on two real world data sets from very different application domains: scene recognition in images and ranking of web-search documents. Its accuracy matches that of the unconstrained baseline (with unlimited resources) while achieving an order of magnitude reduction of test-time cost. Because of its simplicity, high accuracy and drastic test-time cost-reduction we believe our approach to be of strong practical value for a wide range of problems.

## 2. Related Work

Previous work on cost-sensitive learning appears in the context of many different applications. Most prominently, Viola & Jones (2002) greedily train a cascade of weak classifiers with Adaboost (Schapire, 1999) for visual object recognition. Cambazoglu et al. (2010) propose a cascade framework explicitly for web-search ranking. They learn a set of additive weak classifiers using gradient boosting, and remove data points during test-time using proximity scores. Although their algorithm requires almost no extra training cost, the improvement is typically limited. Lefakis & Fleuret (2010) and Dundar & Bi (2007) learn a soft-cascade, which re-weights inputs based on their probability of passing all stages. Different from our method, they employ a global probabilistic model, do not explicitly incorporate feature extraction costs and are restricted to binary classification problems. Saberian & Vasconcelos (2010) also learn classifier cascades. In contrast to prior work, they learn all cascades levels simultaneously in a greedy fashion. Unlike our approach, all of these algorithms focus on learning of cascades and none explicitly focus on individual feature costs.

To consider the feature cost, Gao & Koller (2011) published an algorithm to dynamically extract features during test-time. Raykar et al. (2010) learn classifier cascades, but they group features by their costs and restrict classifiers at each stage to only use a small subset. Pujara et al. (2011) suggest the use of sampling to derive a cascade of classifiers with increasing cost for email spam filtering. Most recently, Chen et al. (2012) introduce *Cronus*, which explicitly considers the feature extraction cost during training and constructs a cascade to encourage removal of unpromising data points early-on. At each stage, they optimize the coefficients of the weak classifiers to minimize the classification error and trees/features extraction costs. We pursue a very different (orthogonal) approach and do not optimize the cascade stages globally. Instead, we strictly incorporate the feature cost into the weak learners. Moreover, as our algorithm is a variant of stage-wise regression, it can operate naturally in both regression and multi-class classification scenarios. (Simultaneous with this publication, Grubb & Bagnell (2012) also proposed a complementary approach to incorporate feature cost into gradient boosting.)

## 3. Notation and Setup

Our training data consist of $n$ input vectors $\{\mathbf{x}_1, \ldots, \mathbf{x}_n\} \in \mathcal{R}^d$ with corresponding labels $\{y_1, \ldots, y_n\} \in \mathcal{Y}$ drawn from an unknown distribution $\mathcal{D}$. Labels can be continuous (regression) or categorial (binary or multi-class classification). We assume that each feature $\alpha$ has an acquisition cost $c_\alpha > 0$ during its initial retrieval. Once a feature has been acquired its subsequent retrieval is free (or set to a small constant).

Further, we are provided an arbitrary continuous loss function $\ell$ and aim to learn a linear predictor $H_{\boldsymbol{\beta}}(\mathbf{x}) = \boldsymbol{\beta}^\top \mathbf{h}(\mathbf{x})$ to minimize the loss function,

$$\min_{\boldsymbol{\beta}} \ell(\boldsymbol{\beta}), \qquad (1)$$

within some test-time cost budget, which will be defined in the following section. One example for $\ell$ is the squared-loss

$$\ell_{sq}(\boldsymbol{\beta}) = \frac{1}{2n} \sum_{i=1}^{n} \left(H_{\boldsymbol{\beta}}(\mathbf{x}_i) - y_i\right)^2, \qquad (2)$$

but other losses, for example the multi-class log-loss (Hastie et al., 2009), are equally suitable. The mapping $\mathbf{x} \to \mathbf{h}(\mathbf{x})$ is a non-linear transformation of the input data that allows the linear classifier to produce non-linear decision boundaries in the original input space. Typically, the mapping $\mathbf{h}$ can be performed *implicitly* through the kernel-trick (Schölkopf, 2001) or *explicitly* through, for example, the boosting-trick (Friedman, 2001; Rosset et al., 2004; Chapelle et al., 2010). In this paper we use the latter approach with limited-depth regression trees (Breiman, 1984). More precisely, $\mathbf{h}(\mathbf{x}_i) = [h_1(\mathbf{x}_i), \ldots, h_T(\mathbf{x}_i)]^\top$, $h_t \in \mathcal{H}$ where $\mathcal{H}$ is the set of all possible regression trees of some limited depth $b$ (e.g. $b = 4$) and $T = |\mathcal{H}|$. The resulting feature space is extremely high dimensional and the weight-vector $\boldsymbol{\beta}$ is always kept to be correspondingly sparse. Because regression trees are negation closed (*i.e.* for each $h \in \mathcal{H}$ we also have $-h \in \mathcal{H}$) we assume throughout this paper w.l.o.g. that $\boldsymbol{\beta} \geq 0$.

Finally, we define a binary matrix $\mathbf{F} \in \{0,1\}^{d \times T}$ in which an entry $F_{\alpha t} = 1$ if and only if the regression tree $h_t \in \mathcal{H}$ splits on feature $\alpha$ somewhere within its tree.

## 4. Method

In this section, we formalize the optimization problem of test-time computational cost, and then intuitively state our algorithm. We follow the setup introduced in (Chen et al., 2012), formalizing the test-time computational cost of evaluating the classifier $H$ for a given weight-vector $\boldsymbol{\beta}$.

**Test-time computational cost.** There are two factors that contribute to this cost: The function evaluation cost of all trees $h_t$ with $\boldsymbol{\beta}_t > 0$ and the feature extraction cost for all features that are used in these trees. Let $e > 0$ be the cost to evaluate one tree $h_t$ if all features were previously extracted. With this notation, both costs can be expressed in a single function as

$$c(\boldsymbol{\beta}) = e\|\boldsymbol{\beta}\|_0 + \sum_{\alpha=1}^{d} c_\alpha \left\| \sum_{t=1}^{T} F_{\alpha t} \beta_t \right\|_0, \qquad (3)$$

where the $l_0$-norm for scalars is defined as $\|a\|_0 \to \{0, 1\}$ with $\|a\|_0 = 1$ if and only if $a \neq 0$. The first term captures the function-evaluation costs and the second term captures the feature costs of all used features. If we combine (1) with (3) we obtain our overall optimization problem

$$\min_{\boldsymbol{\beta}} \ell(\boldsymbol{\beta}), \text{ subject to: } c(\boldsymbol{\beta}) \leq B, \quad (4)$$

where $B \geq 0$ denotes some pre-defined budget that cannot be exceeded during test-time.

**Algorithm.** In the remainder of this paper we derive an algorithm to approximately minimize (4). For better clarity, we first give an intuitive overview of the resulting method in this paragraph. Our algorithm is based on stage-wise regression, which learns an additive classifier $H_{\boldsymbol{\beta}}(\mathbf{x}) = \sum_{t=1}^{m} \beta_t h_t(\mathbf{x})$ that aims to minimize the loss function (4).[1] During iteration $t$, the greedy Classification and Regression Tree (CART) algorithm (Breiman, 1984) is used to generate a new tree $h_t$, which is added to the classifier $H_{\boldsymbol{\beta}}$.

Specifically, CART generates a limited-depth regression tree $h_t \in \mathcal{H}$ by greedily minimizing an impurity function, $g : \mathcal{H} \to \mathcal{R}_0^+$. Typical choices for $g$ are the squared loss (2) or the label entropy (Hastie et al., 2009). CART minimizes the impurity function $g$ by recursively splitting the data set on a single feature per tree-node. We propose an impurity function which on the one hand approximates the negative gradient of $\ell$ with the squared-loss, such that adding the resulting tree $h_t$ minimizes $\ell$, and on the other hand penalizes the initial extraction of features by their cost $c_\alpha$. To capture this initial extraction cost, we define an auxiliary variable $\phi_\alpha \in \{0, 1\}$ indicating if feature $\alpha$ has already been extracted ($\phi_\alpha = 0$) in previous trees, or not ($\phi_\alpha = 1$). We update the vector $\boldsymbol{\phi}$ after generating each tree, setting the corresponding entry for used features $\alpha$ to $\phi_\alpha := 0$. Our impurity function in iteration $t$ becomes

$$g(h_t) = \frac{1}{2} \sum_i \left( -\frac{\partial \ell}{\partial H(x_i)} - h_t(x_i) \right)^2 + \lambda \sum_{\alpha=1}^{d} \phi_\alpha c_\alpha F_{\alpha t}, \quad (5)$$

where $\lambda$ trades off the loss with the cost.

To combine the trees $h_t$ into a final classifier $H_{\boldsymbol{\beta}}$, our algorithm follows the steps of regular stage-wise regression with a fixed step-size $\eta > 0$. As our algorithm is based on a *greedy* opti*miser*, and is stingy with respect to feature-extraction, we refer to it as *the Greedy Miser* (short *miser*). Algorithm (1) shows a pseudo-code implementation.

---
[1]Here, w.l.o.g. the trees in $\mathcal{H}$ are conveniently re-ordered such that exactly the first $m$ trees have non-zero weight $\beta_t$.

**Algorithm 1** *Greedy Miser* in pseudo-code
**Require:** $D = \{(\mathbf{x}_i, y_i)\}_{i=1}^{n}$, step-size $\eta$, iterations $m$
  $H = 0$
  **for** $t = 1$ to $m$ **do**
    $h_t \leftarrow$ Use CART to greedily minimize (5).
    $H \leftarrow H + \eta h_t$.
    For each feature $\alpha$ used in $h_t$, set $\phi_\alpha \leftarrow 0$.
  **end for**
  Return $H$

## 5. Algorithm Derivation

In this section, we derive a connection between (4) and our *miser* algorithm by showing that *miser* approximately solves a relaxed version of the optimization problem.

### 5.1. Relaxation

The optimization as stated in eq. (4) is non-continuous, because of the $l_0$-norm in the cost term—and hard to optimize. We start by introducing minor relaxations to both terms in (3) to make it better behaved.

**Assumptions.** Our optimization algorithm (for details see section 5.2) performs coordinate descent and — starting from $\boldsymbol{\beta} = \mathbf{0}$ — increments one dimension of $\boldsymbol{\beta}$ by $\eta > 0$ in each iteration. Because of the extremely high dimensionality (which is dictated by the number of all possible regression trees that can be represented within the accuracy of the computer) and the comparably tiny number of iterations ($\leq 5000$) it is reasonable to assume that one dimension is never incremented twice. In other words, the weight vector $\boldsymbol{\beta}$ is extremely sparse and (up to re-scaling by $\frac{1}{\eta}$) binary: $\frac{1}{\eta}\boldsymbol{\beta} \in \{0, 1\}^T$.

**Tree-evaluation cost.** The $l_0$-norm is often relaxed into the convex and continuous $l_1$-norm. In our scenario, this is particularly attractive, because if $\frac{1}{\eta}\boldsymbol{\beta}$ is binary, then the re-scaled $l_1$ norm is identical to the $l_0$ norm—and the relaxation is exact. We use this approach for the first term:

$$e\|\boldsymbol{\beta}\|_0 \longrightarrow \frac{e}{\eta}\|\boldsymbol{\beta}\|_1. \quad (6)$$

**Feature cost.** In the case of the feature cost, the $l_1$ norm is not a good approximation of the original $l_0$-norm, because features are re-used many times, in different trees. Using the $l_1$-norm would imply that features that are used more often would be penalized more than features that are only used once. This contradicts our assumption that features become *free* after their initial construction.

We therefore define a new function $q$, which is a re-scaled and amputated version of the $\ell_1$-norm:

$$q(x) = \begin{cases} |\frac{x}{\eta}| & \text{for } |x| \in [0, \eta) \\ 1 & \text{for } |x| \in [\eta, \infty). \end{cases} \quad (7)$$

This penalty function $q$ behaves like the regular $\ell_1$ norm when $|x|$ is small, but is capped to a constant when $x \geq \eta$. With this definition, our relaxation of the feature-cost term becomes:

$$\sum_{\alpha=1}^d c_\alpha \left\| \sum_{t=1}^T F_{\alpha t} \beta_t \right\|_0 \longrightarrow \sum_{\alpha=1}^d c_\alpha q \left( \sum_{t=1}^T F_{\alpha t} \beta_t \right). \quad (8)$$

Similar to the previous case, if $\frac{1}{\eta} \boldsymbol{\beta}$ is binary, this relaxation is exact. This holds because in (8) all arguments of $q$ are non-negative multiples of $\eta$ (as $F_{\alpha t} \in \{0, 1\}$ and $\beta_t \in \{0, \eta\}$) and it is easy to see from the definition of $q$ that for all $k = 0, 1, \ldots$, we have $q(k\eta) = \|k\eta\|_0$.

**Continuous cost-term.** To simplify the optimization, we split the budget into two terms $B = B_t + B_f$—the tree-evaluation budget and the feature extraction budget—and re-write (4) with the two penalties (6) and (8) as two individual constraints. If we use the Lagrangian formulation, with Lagrange multiplier $\lambda$ (up to re-scaling), for the feature cost constraint and the explicit constraint formulation for the tree-evaluation cost, we obtain our final optimization problem:

$$\min_{\boldsymbol{\beta}} \; \ell(\boldsymbol{\beta}) + \lambda \sum_{\alpha=1}^d c_\alpha q \left( \sum_t F_{\alpha t} \beta_t \right) \quad (9)$$

$$s.t. \; \frac{1}{\eta}\|\boldsymbol{\beta}\|_1 \leq \frac{B_t}{e}.$$

### 5.2. Optimization

In this section we describe how *miser*, our adaptation of stage-wise regression (Friedman, 2001), finds a (local) solution to the optimization problem in (9).

**Solution path.** We follow the approach from Rosset et al. (2004) and find a solution path for (9) for evenly spaced tree-evaluation budgets, ranging from $B_t' = 0$ to $B_t' = B_t$. Along the path we iteratively increment $B_t'$ by $\eta$. We repeatedly solve the intermediate optimization problem by warm-starting (9) with the previous solution and allowing the weight vector to change by $\eta$,

$$\min_{\boldsymbol{\delta} \geq 0} \; \overbrace{\ell(\boldsymbol{\beta} + \boldsymbol{\delta}) + \lambda \sum_{\alpha=1}^d c_\alpha q \left( \sum_t F_{\alpha t} (\beta_t + \delta_t) \right)}^{\mathcal{L}(\boldsymbol{\beta}+\boldsymbol{\delta})}, \quad (10)$$

$$s.t. \; \|\boldsymbol{\delta}\|_1 \leq \eta.$$

Each iteration, we update the weight vector $\boldsymbol{\beta} := \boldsymbol{\beta} + \boldsymbol{\delta}$.

**Taylor approximation.** The Taylor expansion of $\mathcal{L}$ is defined as

$$\mathcal{L}(\boldsymbol{\beta} + \boldsymbol{\delta}) = \mathcal{L}(\boldsymbol{\beta}) + \langle \nabla \mathcal{L}(\boldsymbol{\beta}), \boldsymbol{\delta} \rangle + O(\boldsymbol{\delta}^2). \quad (11)$$

If $\eta$ is sufficiently small[2], and because $|\boldsymbol{\delta}| \leq \eta$, we can use the dominating linear term in (11) to approximate the optimization in (10) as

$$\min_{\boldsymbol{\delta} \geq 0} \langle \nabla \mathcal{L}(\boldsymbol{\beta}), \boldsymbol{\delta} \rangle, \; s.t. \; \|\boldsymbol{\delta}\|_1 \leq \eta. \quad (12)$$

**Coordinate descent.** The optimization (12) can be reduced to identifying the direction of steepest descent. Let $\nabla \mathcal{L}(\beta)_t$ denote the gradient w.r.t. the $t^{th}$ dimension, and let us define

$$t^* = \arg\min_t \nabla \mathcal{L}(\beta)_t, \quad (13)$$

to be the gradient dimension of steepest descent. Because $\mathcal{H}$ is negation closed, we have $\nabla \mathcal{L}(\boldsymbol{\beta})_{t^*} = -\|\nabla \mathcal{L}(\boldsymbol{\beta})\|_\infty$. (If $\nabla \mathcal{L}(\boldsymbol{\beta})_{t^*} = 0$ we are done, so we focus on the case when it is $< 0$.) With Hölder's inequality we can derive the following lower bound of the inner product in (12),

$$\langle \nabla \mathcal{L}(\boldsymbol{\beta}), \boldsymbol{\delta} \rangle \geq -|\langle \nabla \mathcal{L}(\boldsymbol{\beta}), \boldsymbol{\delta} \rangle|$$
$$\geq -\|\nabla \mathcal{L}(\boldsymbol{\beta})\|_\infty \|\boldsymbol{\delta}\|_1$$
$$\geq \eta \nabla \mathcal{L}(\beta)_{t^*}. \quad (14)$$

We can now construct a vector $\boldsymbol{\delta}^*$ for which (14) holds as *equality*, which implies that it must be the optimal solution to (12). This is the case if we set $\delta^*_{t^*} = \eta$ and $\delta^*_{\neq t^*} = 0$. Consequently, we can find the solution path with steepest coordinate descent under step-size $\eta$.

**Gradient derivation.** The gradient $\nabla \mathcal{L}(\boldsymbol{\beta})_t$ consists of two parts, the gradient of the loss $\ell$ and the gradient of the feature-cost term. For the latter, we need the gradient of $q\left(\sum_t F_{\alpha t} \beta_t\right)$, which, according to its definition in (7), is not well-defined if $\sum_t F_{\alpha t} \beta_t = \eta$. As our optimization algorithm can only *increase* $\beta_t$, we derive this gradient from the *right*, yielding

$$\nabla q \left( \sum_t F_{\alpha t} \beta_t \right) = \begin{cases} \frac{1}{\eta} F_{\alpha t} & |\sum_t F_{\alpha t} \beta_t| < \eta \\ 0 & |\sum_t F_{\alpha t} \beta_t| \geq \eta. \end{cases} \quad (15)$$

Note that the condition $|\sum_t F_{\alpha t} \beta_t| < \eta$ is true if and only if feature $\alpha$ is not used in any trees with $\beta_t > 0$. Let us define $\phi_\alpha = \{0, 1\}$ with $\phi_\alpha = 1$ iff feature $|\sum_t F_{\alpha t} \beta_t| < \eta$. We can then express the gradient of $\mathcal{L}$ (with a slight abuse of notation) as

$$\nabla \mathcal{L}(\boldsymbol{\beta})_t := \frac{\partial \ell}{\partial \beta_t} + \frac{\lambda}{\eta} \sum_{\alpha=1}^d c_\alpha \phi_\alpha F_{\alpha t}. \quad (16)$$

---

[2]Please note that we see this as a true approximation, and do not expect $\eta$ to be infinitesimally small—which would cause the number of steps (and therefore trees) to become too large for practical use.

Applying the chain rule, we can decompose the first term in (16), $\frac{\partial \ell}{\partial \beta_t}$, into two parts: the derivatives w.r.t. the current prediction $H_{\boldsymbol{\beta}}(\mathbf{x}_i)$, and the partial derivatives of $H_{\boldsymbol{\beta}}(\mathbf{x}_i)$ w.r.t. $\beta_t$. This results in

$$\nabla \mathcal{L}(\boldsymbol{\beta})_t = \sum_{i=1}^n \frac{\partial \ell}{\partial H_{\boldsymbol{\beta}}(\mathbf{x}_i)} \frac{\partial H_{\boldsymbol{\beta}}(\mathbf{x}_i)}{\partial \beta_t} + \frac{\lambda}{\eta} \sum_{\alpha=1}^d c_\alpha \phi_\alpha F_{\alpha t}. \quad (17)$$

As $H_{\boldsymbol{\beta}}(\mathbf{x}_i) = \boldsymbol{\beta}^\top \mathbf{h}(\mathbf{x}_i)$ is linear, we have $\frac{\partial H_{\boldsymbol{\beta}}(\mathbf{x}_i)}{\partial \beta_t} = h_t(\mathbf{x}_i)$. If we define $r_i = -\frac{\partial \ell}{\partial H_{\boldsymbol{\beta}}(\mathbf{x}_i)}$, which we can easily compute for every $\mathbf{x}_i$, we can re-phrase (17) as

$$\nabla \mathcal{L}(\boldsymbol{\beta})_t = \sum_{i=1}^n -r_i h_t(\mathbf{x}_i) + \frac{\lambda}{\eta} \sum_{\alpha=1}^d c_\alpha \phi_\alpha F_{\alpha t}. \quad (18)$$

**The *Greedy Miser*.** For simplicity, we restrict $\mathcal{H}$ to only normalized regression-trees (*i.e.* $\sum_i h_t^2(\mathbf{x}_i) = 1$), which allows us to add two constant terms $\frac{1}{2} \sum_i h_t^2(\mathbf{x}_i)$ and $r_i^2$ to (18) without affecting the outcome of the minimization in (13), as both are independent of $t$. This completes the binomial equation and we obtain a quadratic form:

$$h_t = \operatorname*{argmin}_{h_t \in \mathcal{H}} \frac{1}{2} \sum_i^n (r_i - h_t(x_i))^2 + \lambda' \sum_{\alpha=1}^d c_\alpha \phi_\alpha F_{\alpha t}, \quad (19)$$

with $\lambda' = \frac{\lambda}{\eta}$. Note that (19) is exactly what *miser* minimizes in (5), which concludes our derivation.

**Meta-parameters.** The meta-parameters of *miser* are surprisingly intuitive. The maximum number of iterations, $m$, is tightly linked to the tree-evaluation budget $B_t$. The optimal solution of (12) must satisfy the *equality* $\|\boldsymbol{\delta}^*\|_1 = \eta$ (unless $\nabla \mathcal{L} = \mathbf{0}$, in which case a local minimum has been reached and the algorithm would terminate). As $\|\boldsymbol{\beta}\|_1$ is exactly increased by $\eta$ in each iteration, it can be expressed in terms of the number of iterations $m$ of the algorithm, and we obtain $\frac{1}{\eta}\|\boldsymbol{\beta}\|_1 = m$. Consequently, in order to satisfy the $l_1$ constraint in (9), we must limit to the number of iterations to $m \leq \frac{B_t}{e}$. The parameter $\lambda'$ corresponds directly to the feature-budget $B_f$. The algorithm is not particularly sensitive to the exact step-size $\eta$, and throughout this paper we set it to $\eta = 0.1$.

## 6. Results

We conduct experiments on two benchmark tasks from very different domains: the Yahoo Learning to Rank Challenge data set (Chapelle & Chang, 2011) and the scene recognition data set from Lazebnik et al. (2006).

**Yahoo Learning to Rank.** The Yahoo data set contains document/query pairs with label values from $\{0, 1, 2, 3, 4\}$, where 0 means the document is irrelevant to the query, and 4 means highly relevant. In total, it has 473134, 71083, 165660, training, validation, and testing pairs. As this is a regression task, we use the squared-loss as our loss function $\ell$. Although the data set is representative for a web-search ranking *training* data set, in a real world *test* setting, there are many more irrelevant data points. Usually, for each query, only a few documents are relevant, and the other hundreds of thousands are completely irrelevant. Therefore, we follow the convention of Chen et al. (2012) and replicate each irrelevant data point (label value is 0) 10 times.

Each feature in the data set has an acquisition cost. The feature costs are discrete values in the set $\{1, 5, 10, 20, 50, 100, 150\}$. The unit of these costs is approximately the time to evaluate a feature. The cheapest features (cost value is 1) are those that can be acquired by looking up a table (such as the statistics of a given document), whereas the most expensive ones (such as BM25F-SD described in Broder et al. (2010)), typically involve term proximity scoring.

To evaluate the performance on this task, we follow the typical convention and use Normalized Discounted Cumulative Gain (NDCG@5) (Järvelin & Kekäläinen, 2002), as it places stronger emphasis on retrieving relevant documents within a large set of irrelevant documents.

**Loss/cost trade-off.** Figure 1 (*left*) shows the traces (dashed lines) of the NDCG@5/cost generated by repeatedly adding trees to the predictor until 3000 trees in total — essentially depicting the results under increasing tree-evaluation budgets $B_t$. The different traces are obtained under varying values of the feature-cost trade-off parameter $\lambda$. The baseline, *stage-wise regression* (Friedman, 2001), is equivalent to *miser* with $\lambda = 0$ and is essentially building trees without any cost consideration. The red circles indicate the iteration with the highest NDCG@5 value on the validation data set. The graph shows, that under increased $\lambda$ (the solid red line), the NDCG@5 ranking accuracy of *miser* drops very gradually, while the test-time cost is reduced drastically (compared to $\lambda = 0$).

**Comparison with prior work.** In addition to stage-wise regression, we also compare against *Stage-wise regression feature subsets*, *Early Exit* (Cambazoglu et al., 2010) and *Cronus* (Chen et al., 2012). *Stage-wise regression feature subsets* is a natural extension to stage-wise regression. We group all features according to the feature cost, and gradually use more expensive feature groups. The curve is generated by only using features whose cost $\leq 1, 20, 100, 200$. *Early Exit*, proposed by Cambazoglu et al. (2010), trains trees identical to stage-wise regression—however, it reduces the average test-time cost by removing unpromising documents early-on during test-time. Among all methods

The Greedy Miser

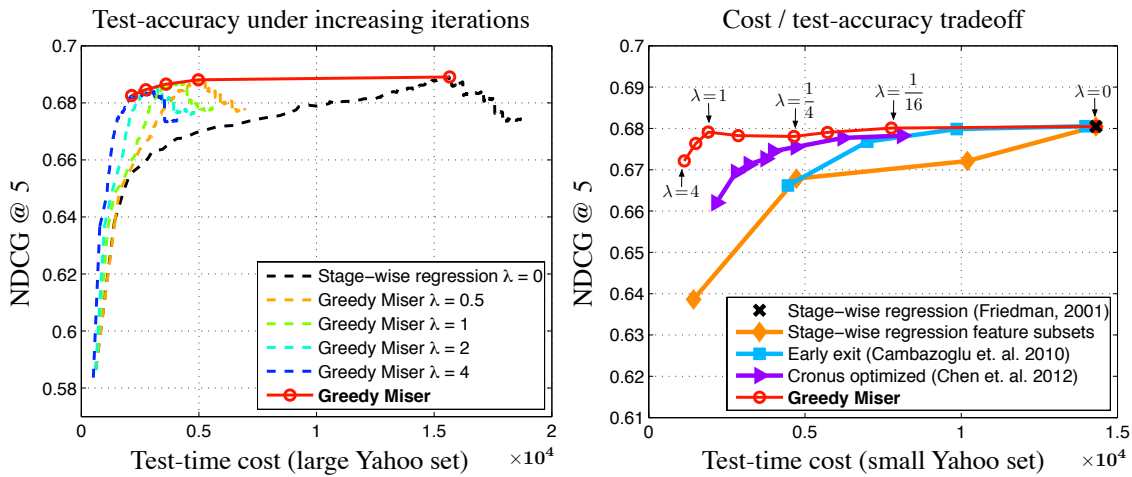

*Figure 1.* The NDCG@5 and the test-time cost of various classifier settings. *Left:* The comparison of the original *Stage-wise regression* ($\lambda = 0$) and *miser* under various feature-cost/accuracy trade-off settings ($\lambda$) on the full Yahoo set. The dashed lines represent the NDCG@5 as trees are added to the classifier. The red circles indicate the best scoring iteration on the validation data set. *Right:* Comparisons with prior work on test-time optimized cascades on the small Yahoo set. The cost-efficiency curve of *miser* is consistently above prior work, reducing the cost, at similar ranking accuracy, by a factor of 10.

of early-exit the authors suggested, we plot the best performing one (Early Exit Using Proximity Threshold). We introduce an early exit every 10 trees (300 in total), and at the $i^{th}$ early-exit, we remove all test-inputs that have a score of at least $\frac{(300-i)s}{299}$ lower than the fifth best input (where $s$ is a parameter regulating the pruning aggressiveness). The overall improvement over stage-wise regression is limited because the cost is dominated by the feature acquisition, rather than tree computation. It is worth pointing out that the cascade-based approaches of Early-Exits and Cronus are actually complementary to *miser* and future work should combined them.

Since *Cronus* does not scale to the full data set, we use the subset of the Yahoo data from Chen et al. (2012) of 141397, 146769, 184968, training, validation and testing points respectively. In comparison to *Cronus*, which requires $O(mn)$ memory, *miser* requires no significant operational memory besides the data and scales easily to millions of data points. Figure 1 (*right*) depicts the trade-off curves, of *miser* and competing algorithms, between the test-time cost and generalization error. We generate the curves by varying the feature-cost trade-off $\lambda$ (or the pruning parameter $s$ for *Early-Exits*). For each setting we choose the iteration that has the best validation NDCG@5 score. The graph shows that all algorithms manage to match the unconstrained cost-results of stage-wise regression. However, the trade-off curve of *miser* stays consistently above that of *Cronus* and *Early Exits*, leading to better ranking accuracy at lower test-time cost. In fact, *miser* can almost match the ranking accuracy of stage-wise regression with 1/10 of the cost, whereas *Cronus* reduces the cost only to 1/4 and *Early-Exits* to 1/2.

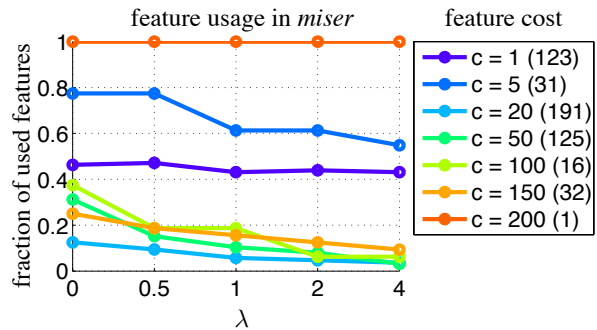

*Figure 2.* Features (grouped by cost $c$) used in *miser* with various $\lambda$ (the number of features in each cost group is indicated in parentheses in the legend). Most cheap features ($c = 1$) are extracted constantly in different $\lambda$ settings, whereas expensive features ($c \geq 5$) are extracted more often when $\lambda$ is small. The most expensive (and invaluable) feature $c = 200$ is always extracted.

**Feature extraction.** To investigate what effect the feature-cost trade-off parameter $\lambda$ has on the classifier's feature choices, Figure 2 visualizes what type of features are extracted by *miser* as $\lambda$ increases. For this visualization, we group features by cost and show what fraction of features in each group are extracted. The legend in the right indicates the cost of a feature group and the number of features that fall into it (in the parentheses). We plot the feature fraction at the best performing iteration based on the validation set. With $\lambda = 0$, *miser* does not consider the feature cost when building trees, and thus extracts a variety of expensive features. As $\lambda$ increases, it extracts fewer expensive features and re-uses more cheap features ($c_\alpha = 1$). It is interesting to point out that across all different *miser* settings, a few expensive features (cost≥150) are always extracted within early iterations. This highlights a great advantage of *miser* over some other cascade algorithms (Raykar et al., 2010), which learn cascades with pre-assigned feature costs and cannot extract good but expensive features until the very end.

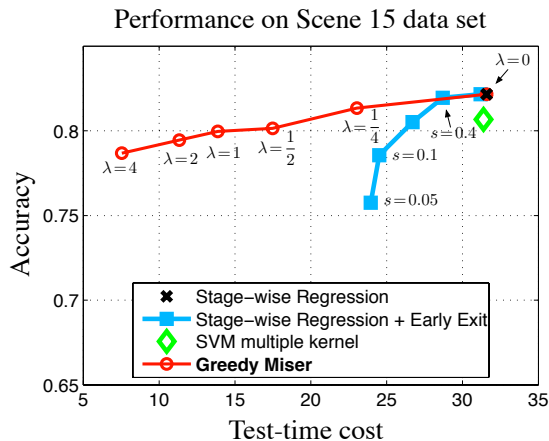

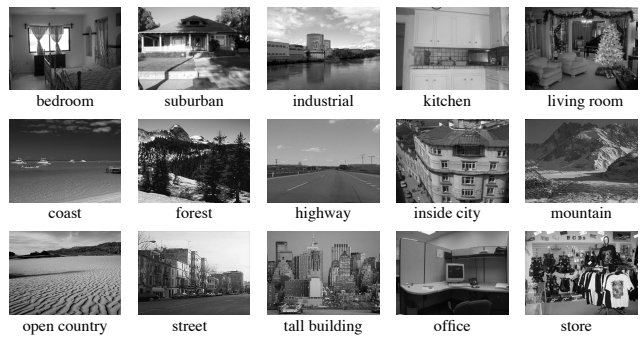

Figure 4. Sample images of the Scene 15 classification task.

Figure 3. Accuracy as a function of cpu-cost during test-time. The curve is generated by gradually increasing $\lambda$. *Miser* champions the accuracy/cost tradeoff and obtains similar accuracy as the SVM with multiple kernels with only half its test-time cost.

**Scene Recognition.** The Scene-15 data set (Lazebnik et al., 2006) is from a very different data domain. It contains 4485 images from 15 scene classes and the task is to classify images according to scene. Figure 4 shows one example image for each scene category. We follow the procedure used by Lazebnik et al. (2006); Li et al. (2010), randomly sampling 100 images from each class, resulting in 1500 training images. From the remaining 2985 images, we randomly sample 20 images from each class as validation, and leave the rest 2685 for test.

We use a diverse set of visual descriptors varying in computation time and accuracy: GIST, spatial HOG, Local Binary Pattern, self-similarity, texton histogram, geometric texton, geometric texton, geometric color, and Object Bank (Li et al., 2010). The authors from Object Bank apply 177 object detectors to each image, where each object detector works independently of each other. We treat each object detector as an independent descriptor and end up with a total of 184 different visual descriptors.

We split the training data 30/70 and use the smaller subset to construct a kernel and train 15 one-vs-all SVMs for each descriptor. We use the predictions of these SVMs on the larger subset as the features of *miser* (totaling $d=184\times15=2760$ features.) As loss function $\ell$, we use the multi-class log-loss (Hastie et al., 2009) and maintain 15 tree-ensemble classifiers $H^1, \ldots, H^{15}$, one for each class. During each iteration, we construct 15 regression trees (depth 3) and update all classifiers. For a given image, each classifier's (normalized) output represents the probability of this data point belonging to one class.

We compute the feature-extraction-cost as the cpu-time required for the computation for the visual descriptor, the kernel construction and the SVM evaluation. Each visual descriptor is used by 15 one-vs-all features. The moment any one of these features is used, we set the feature extraction cost of all other features that are based on the same visual descriptor to only the SVM evaluation time (*e.g.* if the first HOG-based feature is used, the cost of all other HOG-based features is reduced to the time required to evaluate the SVM). Figure 3 summarizes the results on the Scene-15 data set. As baseline we use stage-wise regression (Friedman, 2001) and an SVM with the averaged kernel of all descriptors. We also apply stage-wise regression with *Early Exits*. As this is multi-class classification instead of regression we introduce an early exit every 10 trees (300 in total), and we remove test-inputs whose maximum class-likelihood is greater than a threshold $s$. We generate the curve of early exit by gradually increasing the value for $s$. The last baseline is original vision features with $\ell_1$ regularization, and we notice that its accuracy never exceeds 0.74, and therefore we do not plot it. The *miser* curve is generated by varying loss/feature-cost trade-off $\lambda$. For each setting we choose the iteration that has the best validation accuracy, and all results are obtained by averaging over 10 randomly generated training/testing splits.

Both, multiple-kernel SVM and stage-wise regression achieve high accuracy, but their need to extract all features significantly increases their cost. Early Exit has only limited improvement due to the inability to select a few expensive but important features in early iterations. As before, *miser* champions the cost/accuracy trade-off and its accuracy drops gently with increasing $\lambda$.

All experiments (on both data sets) were conducted on a desktop with dual 6-core Intel i7 cpus with 2.66GHz. The training time for *miser* requires comparable amount of time as stage-wise regression (about 80 minutes for the full Yahoo data set and 12 minutes for Scene-15.)

## 7. Conclusion

Accounting for the operational cost of machine learning algorithms is a crucial problem that appears throughout current and potential applications of machine learning. We believe that understanding and controlling this trade-off will become a fundamental part of machine-learning research in the near future. This paper introduces a natural extension to stage-wise regression (Friedman, 2001), which incorporates feature cost during training. The resulting algorithm, the *Greedy Miser*, is simple to implement, naturally scales to large data sets and outperforms previously most cost-effective classifiers.

Future work includes combining our approach with *Early Exits* (Cambazoglu et al., 2010) or cascade based learning methods such as (Chen et al., 2012).

## 8. Acknowledgements

KQW and ZX would like to thank NIH for their support through grant U01 1U01NS073457-01.